\documentclass{article}

\pdfoutput=1
\usepackage{fullpage}
\usepackage[utf8]{inputenc}
\usepackage{graphicx}
\usepackage{amssymb}
\usepackage{amsmath}
\usepackage{amsfonts}
\usepackage{tabularx,colortbl}
\usepackage{multirow}
\usepackage{color}
\usepackage[pdftex]{hyperref}
\usepackage[english]{babel}
\usepackage{subfigure}
\usepackage{algorithm}
\usepackage{algorithmic}
\usepackage[numbers,sort&compress]{natbib}
\usepackage{authblk}

\makeindex

\definecolor{Orange}{rgb}{1,0.64,0}
\definecolor{lgray}{rgb}{0.85,0.85,0.85}

\begin{document}

\title{Classifying sequences by the optimized dissimilarity space embedding approach: a case study on the solubility analysis of the {E}. coli proteome}

\author[1]{Lorenzo Livi\thanks{llivi@scs.ryerson.ca}\thanks{Corresponding author}}
\author[2]{Antonello Rizzi\thanks{antonello.rizzi@uniroma1.it}}
\author[1]{Alireza Sadeghian\thanks{asadeghi@ryerson.ca}}
\affil[1]{Dept. of Computer Science, Ryerson University, 350 Victoria Street, Toronto, ON M5B 2K3, Canada}
\affil[2]{Department of Information Engineering, Electronics, and Telecommunications, SAPIENZA University of Rome, Via Eudossiana 18, 00184 Rome, Italy}
\renewcommand\Authands{, and }
\providecommand{\keywords}[1]{\textbf{\textit{Index terms---}} #1}

\maketitle

\begin{abstract}
We evaluate a version of the recently-proposed classification system named Optimized Dissimilarity Space Embedding (ODSE) that operates in the input space of sequences of generic objects. The ODSE system has been originally presented as a classification system for patterns represented as labeled graphs. However, since ODSE is founded on the dissimilarity space representation of the input data, the classifier can be easily adapted to any input domain where it is possible to define a meaningful dissimilarity measure.
Here we demonstrate the effectiveness of the ODSE classifier for sequences by considering an application dealing with the recognition of the solubility degree of the Escherichia coli proteome.
Solubility, or analogously aggregation propensity, is an important property of protein molecules, which is intimately related to the mechanisms underlying the chemico-physical process of folding.
Each protein of our dataset is initially associated with a solubility degree and it is represented as a sequence of symbols, denoting the 20 amino acid residues.
The herein obtained computational results, which we stress that have been achieved with no context-dependent tuning of the ODSE system, confirm the validity and generality of the ODSE-based approach for structured data classification.\\
\end{abstract}

\maketitle


\section{Introduction}

A considerable number of pattern recognition applications deal with data represented as sequences of objects \cite{xing2010,Bicego20042281}.
The handwriting on-line recognition problem provides a good example: each object in the sequence is a distinct stroke written by a user through a suitable input device \cite{10.1109/34.57669}.
The objects of the sequence can originate from a finite alphabet, as in the problem of text excerpt classification \cite{4509437}, where the objects are usually words of the corresponding language. Another interesting example pertains biochemical compounds, such as DNA and proteins \cite{smialowski2007protein,giuliani2002,xiaohui2014predicting}.
Many applications deal with objects originating from continue domains, such as in the case of series of real-valued data (e.g., 3D spatial positions of a moving agent, financial time-series, and sequences of measurements related to a physical plant or device) \cite{10.1109/TKDE.2012.223,pedrycz_timeseries}.
Often the ordering of the objects is assumed to be over the time domain; in this case, the objects are referred to as \textit{events}.
When dealing with classification problems defined on an input space of sequences, modeling the abstract classification function underlying the unknown data generating process can become a very complex task. To this end, the problem can be faced by mapping the sequences into $\mathbb{R}^d$ vectors, using a suitable feature extraction procedure \cite{grapsec_ijcnn_2013,livi+delvescovo+rizzi_seriation+gradis,delvescovo_rlgradis_2012,seriation+gradis_lncs_2012,gralg_2012}.
In this scenario, features are elaborated from the dataset at hand, and therefore can be considered as a \textit{local reference framework} on which ground the new vector-based representation of the input data.
This solution is supported by the fact that there are many effective and well-established data-driven inductive inference systems that deal with real-valued vectors as input patterns \cite{rizzi2002,1009136,melin2013review,melin2014new,sanchez2014optimization,melin2014review}.

The dissimilarity representation is a powerful framework for dealing with pattern recognition and data mining problems defined in spaces with no trivial (geo)metric structure \cite{Bicego20042281,Duin2012826,odse,delvescovo+rizzi2007,cagata,riesen+bunke2010,5597597,Jacobs:2000:CND:355091.355095,trujillo2013parzen}.
Pairwise dissimilarity values of input data are generated according to a suitable dissimilarity measure, $d: \mathcal{X}\times\mathcal{X}\rightarrow\mathbb{R}^+$, which operates directly in the input domain, $\mathcal{X}$. Such a dissimilarity measure, which can also diverge from a common metric distance (e.g., Euclidean distance), is usually defined by parameters that allow to tailor the matching method to the particular data type at hand. The dissimilarity representation is developed with respect to a local base of prototypes, $\mathcal{R}$, called \textit{representation set}.
Dissimilarity-based classification systems for sequences of objects have been proposed on various data types, such as simple atomic elements \cite{grapsec_ijcnn_2013} (e.g., sequences of numbers or characters belonging to a finite alphabet) and more complex data structures, like type-2 fuzzy sets \cite{t2apdiss__ifsanafips2013,it2fs_matching,usit2_2012} and pen strokes \cite{delvescovo07OnlineHandwritingRecognitionbySymbolicHistogramsApproach,batista+granger+sabourin2010,10.1109/34.57669}.

The recently proposed classification system called Optimized Dissimilarity Space Embedding (ODSE) is founded on the dissimilarity representation. It has been originally developed as a classification system for labeled graphs, denoting state-of-the-art test set performances on various well-known benchmarking datasets \cite{odse,odse2_ijcnn_2013}.
The synthesis of the classification model is implemented by exploiting an information-theoretic interpretation of the dissimilarity matrix. Such an interpretation allows to conceive suitable \textit{compression} and \textit{expansion} operations of the representation set, implementing the optimization of the dissimilarity space.
The ODSE system performs all operations (i.e., synthesis of the model, test set evaluation etc.) starting from the construction of the dissimilarity matrix, which in turn is completely elaborated through a dissimilarity measure.
Therefore, by adapting the dissimilarity measure to the particular context at hand, it is possible to easily adapt the whole ODSE system accordingly.

In this paper, we present and evaluate a version of the ODSE system tailored to work in the input space of sequences.
The main aim of this study is to demonstrate the versatility and effectiveness of the design underlying the ODSE classifier, which is quickly adaptable to different input domains and hence application contexts. 
Toward this end, here we evaluate the ODSE classifier for sequences on an important application concerning the solubility analysis of sequences of amino acid residues, elaborated from the Escherichia coli (E. coli) proteome \cite{niwa2009,Niwa05062012}. We contrast our experimental results with two reference systems and with respect to another recently-proposed classifier for sequences \cite{grapsec_ijcnn_2013}.
Folding is a chemico-physical process of extraordinary difficulty and complexity from the viewpoint of prediction.
This fact is due to (i) the large number of residues constituting protein molecules and to (ii) the multiplicity of different energetic constraints involved in the underlying physical process \cite{dill2012protein}.
Moreover, the process of folding is in strict competition with the aggregation process (low propensity of a molecule to be soluble), that is, with the tendency of establishing inter-molecular bonds. This results in the formation of large multi-molecular aggregates which, analogously to what happens for artificial polymers, are insoluble and hence precipitate in solution \cite{dill2012protein,giuliani2002,frauenfelder1994}.

The herein considered data as been already processed by different groups \cite{6404416,grapsec_ijcnn_2013,Agostini2012237}. Therefore, we use our previous results \cite{grapsec_ijcnn_2013} for comparison in the herein presented experiments.

This paper is structured as follows.
Section \ref{sec:odse} quickly introduces the ODSE system; initially we describe the system in terms of graph-based patterns. Section \ref{sec:odse_seq} discusses the straightforward adaptation of the ODSE system for processing sequences.
In Section \ref{sec:exps} we discuss the experiments performed on the E. coli dataset. Finally, Section \ref{sec:conclusions} concludes the paper.

\section{The ODSE Classification System}
\label{sec:odse}

The ODSE graph classification system \cite{odse,odse2_ijcnn_2013} is based on the interplay between different techniques, among which we have graph matching, dissimilarity space representation, cluster analysis, and information-theoretic data analysis methods. The classifier is founded on an explicit graph embedding mechanism that represents the input graphs $\mathcal{S}, n=|\mathcal{S}|$, using a suitable representation set, $\mathcal{R}, m=|\mathcal{R}|$, by computing the dissimilarity matrix $\mathbf{D}^{n\times m}$.
Originally, the system has been conceived to operate on the labeled graphs domain $\mathcal{G}$ by means of a suitable inexact graph matching procedure \cite{gm_survey}.
The vector configuration representing the input data in the embedding space $\mathcal{D}$ is derived directly using the rows of $\mathbf{D}$.

An important component of the ODSE graph classification system is the inner feature-based classifier, which operates directly on the developed dissimilarity space embedding; its own classification model is synthesized along with the ODSE synthesis.
Such a classifier can be any well-known classification system, such as a neuro-fuzzy Min-Max network \cite{rizzi2002,Gabrys2000}.

\subsection{A Quick Look Into the ODSE Design}
\label{sec:odse_design}

\begin{figure*}[ht!]
\centering
\subfigure[ODSE embedding space synthesis step.]{
   \includegraphics[viewport=0 0 662 270,scale=0.38,keepaspectratio=true]{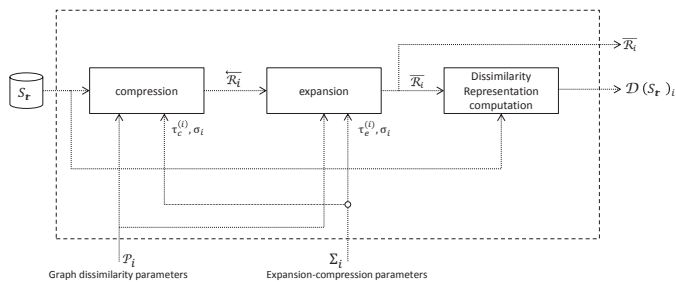}
   \label{fig:odse_emb}
 }
~
 \subfigure[Synthesis of the ODSE classification model.]{
   \includegraphics[viewport=0 0 505 241,scale=0.4,keepaspectratio=true]{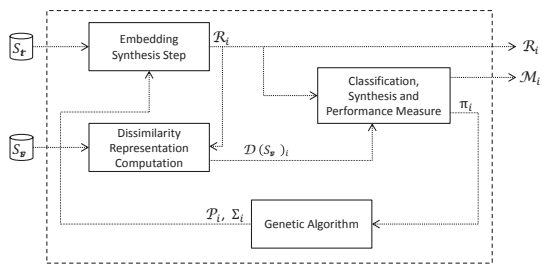}
   \label{fig:odse_synth}
 }
\caption{Schematic description of the ODSE embedding space and classification model synthesis. Taken from \cite{odse}.}
\label{fig:odse}
\end{figure*}

Test patterns are classified by ODSE feeding the corresponding dissimilarity representation to the (already synthesized) feature-based classifier, which assigns proper class labels to the test patterns.
Figs. \ref{fig:odse_emb} and \ref{fig:odse_synth} provide, respectively, the diagrams of the ODSE embedding procedure and of the model synthesis.
The ODSE classification model is defined by the representation set $\mathcal{R}_i$, the setting of the inexact graph matching function parameters (denoted as $\mathcal{P}_i$), and the feature-based classifier (denoted as $\mathcal{M}_i$ in Fig. \ref{fig:odse_synth}) on the developed dissimilarity space.
During the synthesis stage, additional parameters are optimized, which are fundamental to the determination of the \textit{optimal} classification model. Those parameters, which are synthetically denoted as $\Sigma_i$ in Fig. \ref{fig:odse_emb}, are the kernel size $\sigma$ used by the entropy estimator and the two entropy thresholds $\tau_c, \tau_e$, which play a fundamental role in the compression and expansion operations, respectively.
The former is used to reduce the number of prototypes, while the latter replaces targeted prototypes that do not help discriminating the input data represented in the dissimilarity space.
Both operations make use of a suitable non-parametric $\alpha$-order R\'{e}nyi entropy estimator to characterize the informativeness of the prototypes.
Since the dissimilarity values fall into a continuous interval, the underlying distribution is assumed to be continuous as well.
So far, the ODSE system has been tested considering two well-known entropy estimators. The first estimator has been proposed by \citet{principe2010}, and it is referred to as the Quadratic R\'{e}nyi entropy (QRE) estimator, while the second one is based on the construction of the so-called \textit{entropic} Minimum Spanning Tree (MST) among the data samples \cite{gs:HeroEtAl2002}.
While both estimators showed, as ODSE components, comparable performances, the latter is considerably faster and less sensible to the dimensionality of the data (i.e., dissimilarity vectors).

In current implementations \cite{odse,odse2_ijcnn_2013,odse2__arxiv}, the ODSE model is optimized through a genetic algorithm (GA).
The GA operates by performing roulette wheel selection, two-point crossover, and random mutation on the variables representing the model parameters; it implements also an elitism strategy that includes the fittest individual into the next population. The GA, in practice, evolves a population of ODSE model instances that are evaluated by considering a suitable fitness function.
Such a fitness function takes into account a combination of the recognition rate $\pi_i$ achieved on a validation set, $\mathcal{S}_{vs}$, the cardinality of the compressed-and-expanded representation set, and finally the estimated entropy related to the embedded training data.
Convergence criterion is determined as a combination of a maximum number of iterations and a check that evaluates the variation of the fitness in the last five iterations.

\subsection{Classifying Sequences with ODSE}
\label{sec:odse_seq}

It is easy to realize that ODSE can be straightforwardly adapted to operate in other input spaces (i.e., different from the labeled graphs domain).
In fact, once an effective dissimilarity measure $d(\cdot, \cdot)$ that operates in the specific input space at hand (e.g., a domain of sequence of characters) is defined, all ODSE operations are automatically valid and well-defined, since they are performed on the vectors derived from the dissimilarity matrix $\mathbf{D}$ (i.e., compression, expansion, and entropy estimation).
Therefore, adaptation of ODSE to process sequences of generic objects is performed by implementing $d(\cdot, \cdot)$ as a suitable sequence matching algorithm.

From the software implementation point of view, ODSE flexibility is due to the C++ template metaprogramming approach\footnote{\url{http://sourceforge.net/p/libspare/home/Spare/}} \cite{spare_graph_2013}.
In fact, the designer is in charge to define just a class implementing a suitable dissimilarity measure, passing it as an argument to the main procedure.

\section{Recognition of the E. coli Protein Solubility}
\label{sec:exps}

Aggregation propensity of proteins is strongly related to ``errors'' in the folding process \cite{dill2012protein}.
In fact, protein aggregation is at the basis of pathologies defined as \textit{misfolding diseases} \cite{Taubes15031996}, which include Alzheimer and Parkinson. Nevertheless, protein--protein interactions are of vital importance for proteins exerting their physiological role, so that a refined balance between aggregation (inter-molecular connections) and folding (intra-molecular connections) is of utmost importance.
This balance is so crucial for life that does exist a class of protein molecules called ``chaperones'', whose specific role is to help other proteins in completing a correct folding process \cite{shtilerman1999}. However, the nature of the \textit{code} by which a linear sequence of amino acid residues is transformed into a functional 3D structure is still elusive \cite{dill2012protein}.
It is well-known that some proteins are capable to easily reach the stable state, so that they can undergo different folding--unfolding cycles even when isolated from the cell micro-environment.
On the other hand, there are other proteins that cannot be folded when isolated from their cellular environment, necessitating a chaperone-driven folding process.

\citet{niwa2009} studied, in a strictly controlled setting, the aggregation/solubility propensity of the E. coli proteins.
Proteins having difficulty in performing the folding autonomously (i.e., without the help of chaperones) tend to aggregate and hence precipitate in the solution (water in this case).
\citet{Agostini2012237} in a recent work clearly demonstrated that the solubility degree in the same Niwa et al. data base negatively correlates with the aggregation propensity (the aggregation is estimated from the folded state).
This result implies that it is possible to consider the solubility as a measure of the relative stability of both the folded and aggregated states (i.e., the higher the solubility the higher the protein stability in its native state).

\subsection{Dataset Description}
\label{sec:dataset}

\begin{figure*}[ht!]
\centering
\subfigure[Density of the 3173 E. coli proteins with respect to the normalized solubility degree.]{
   \includegraphics[viewport=0 0 341 243,scale=0.6,keepaspectratio=true]{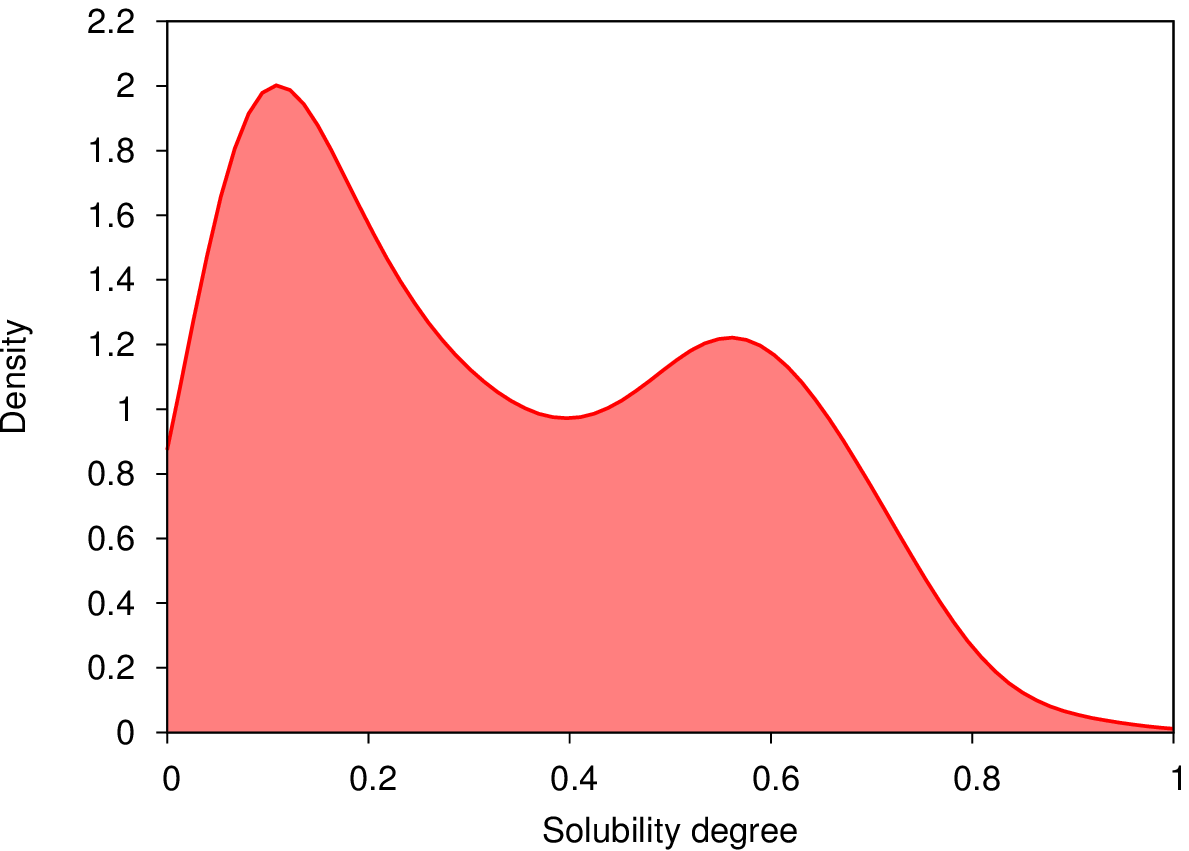}
   \label{fig:solubility}
 }
~
 \subfigure[Normalized solubility degree of the considered 1811 proteins. The two classes are clearly recognized.]{
   \includegraphics[viewport=0 0 351 241,scale=0.6,keepaspectratio=true]{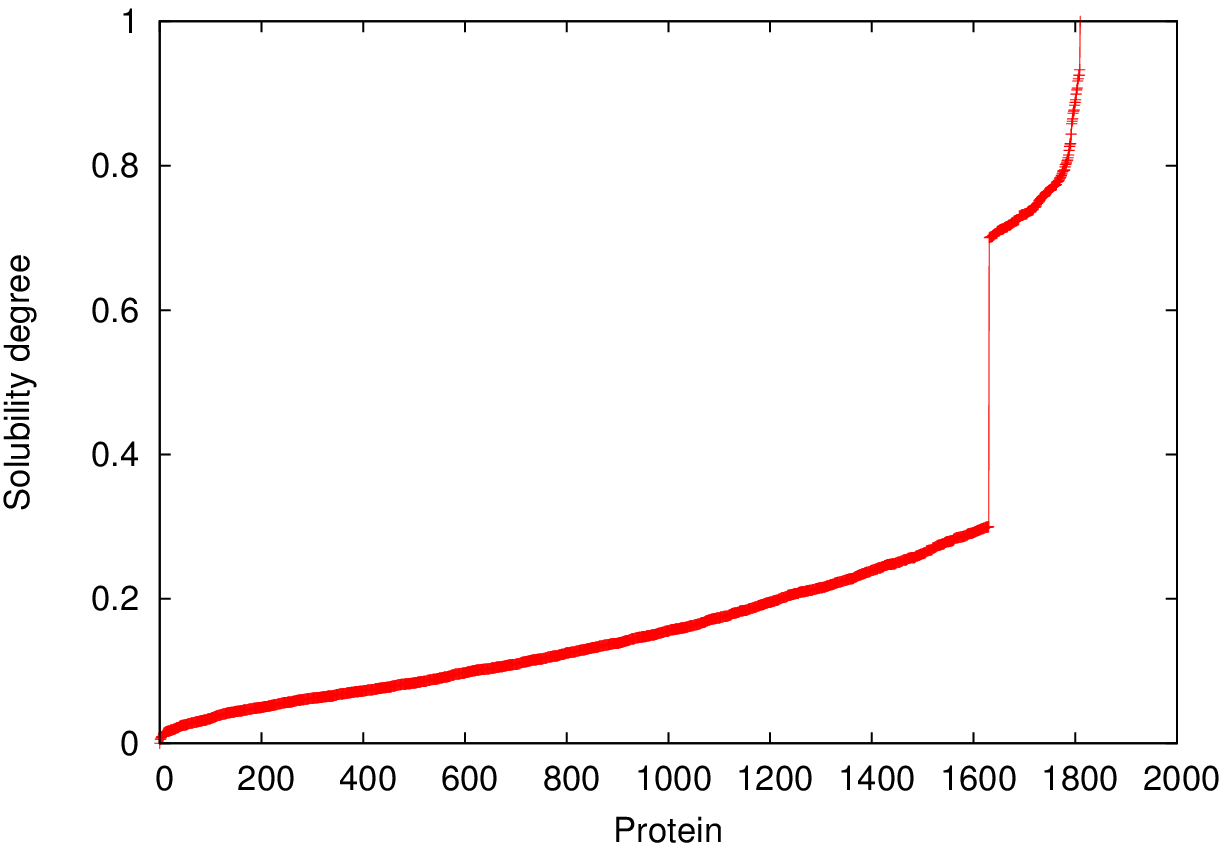}
   \label{fig:norm_sol}
 }
\caption{Density (\ref{fig:solubility}) and solubility degree (\ref{fig:norm_sol}) of the considered proteins.}
\label{fig:figs}
\end{figure*}

The 3173 E. coli proteins elaborated by \citet{niwa2009} were transcribed and translated from the E. coli DNA extracts in strictly controlled conditions.
Their solubility was assessed in terms of percentage of protein concentration at saturation point. The authors demonstrated a bi-modal distribution of solubility, with many poorly soluble proteins and a fairly smaller set of very soluble proteins (see Fig. \ref{fig:solubility}).
For instance, when considering $[0, 0.3]$ and $[0.7, 1]$ as the two intervals of normalized solubility characterizing the insoluble and soluble proteins, the dataset would be split into 1631 insoluble and 180 soluble proteins, respectively, which makes the corresponding classification problem very unbalanced \cite{grapsec_ijcnn_2013}. Those interval of solubility, although they generate an unbalanced classification problem, are of the same length and they are placed at the extremes of the (normalized) solubility range. This fact reassures us to perform a fair (although perhaps non optimal) construction of the soluble and insoluble classes (see Fig. \ref{fig:norm_sol}).
The input of the herein considered classifiers consists in the E. coli sequences of amino acid identifiers, which output the predicted ``soluble'' or ``insoluble'' class.

As a consequence of the aforementioned bi-modality of the solubility degree density, the herein discussed experiments have been organized by considering two different splits of the original dataset.
The first experiment (setting previously considered here \cite{grapsec_ijcnn_2013}) operates over a perfectly balanced (small) dataset made of the 100 proteins with the highest solubility degree and the 100 with the lowest solubility degree. We refer to this dataset as DS-200. The training set $\mathcal{S}_{tr}$ contains 140 proteins and the test set $\mathcal{S}_{ts}$ the remaining 60. Training and test set are characterized by the same number of soluble and insoluble proteins.
The second experiment, which we introduce here, takes into account instead a larger dataset made of 1811 proteins (DS-1811), obtained by considering the aforementioned solubility ranges to define the two classes (see Fig. \ref{fig:norm_sol}). DS-1811 contains a total 179 soluble and 1632 insoluble proteins. 
We consider two different split settings for DS-1811.
In the first case, the training set contains 180 proteins, 70 of which belong to the soluble class and the remaining 110 to the insoluble class. The test set is considerably larger, since in fact it contains 1631 proteins, 1521 of which belong to the insoluble class.
The training set, although it is much smaller than the test set, it is suitably conceived to ``cover'' the considered data with characterizing proteins, that is, with proteins that suitably represent the soluble or insoluble prototypical patterns -- those proteins are selected as prototypes of the respective classes by exploiting a preliminary clustering-based analysis.
In the second case, we consider a perfectly balanced training set, made of 100 soluble and 100 insoluble proteins randomly selected among the 1811 that are available in DS-1811.
The remaining 1611 proteins form the test set. We refer to this dataset instance as DS-1811-2. The motivation behind this further split setting is due to the fact that typically classifiers are sensitive to the percentages of per-class training set patterns. We will asses also this aspect in our experiments.

\subsection{Experimental Setting}

In this paper, we adopt the ODSE version described in \cite[Sec.~3]{odse2_ijcnn_2013}. For the sake of shortness, we do not report here the details of this particular ODSE version, referring the reader to given reference.
We setup ODSE to operate in the input space of sequences by means of a dissimilarity measure implemented through the Levenshtein sequence matching algorithm \cite{t2apdiss__ifsanafips2013,t2vsdiss__ifsanafips2013}. We consider in the Levenshtein global alignment scheme suitable pre-computed \textit{substitution weights}; we used the weights provided by the PAM120 matrix, which has been retrieved from: \url{ftp://ftp.ncbi.nih.gov/blast/matrices/}.
We report the results obtained with two different feature-based classifiers operating in the dissimilarity space: a \textit{k}-NN rule operating the Euclidean distance (kNN) and the C-SVM classifier (C-SVM) equipped with a Gaussian kernel.
We compare ODSE with the recently-proposed sequence classification system based on another embedding technique \cite{grapsec_ijcnn_2013}, which is denoted as GRAPSEC in the following.
We consider also two reference systems that operate directly in the input space. The first one is a \textit{k}-NN rule based classifier (kNN), equipped with the same weighted Levenshtein matching scheme used in ODSE. The second one is a \textit{kernelized} C-SVM classifier operating in the input space of sequences through the kernel function elaborated from the Levenshtein metric -- no corrections are performed to assure positive definiteness of the resulting kernel.
Setting of meta-parameters (e.g., $C$ of C-SVM) have been defined by preliminary tests.
In the following, ``0'' indicates the insoluble class, while ``1'' the soluble class.
To make our results statistically significant, we considered 10 different randomized re-samplings of the dataset DS-1811 (both split settings). Split percentages of both settings are defined as described in the previous section. The following results on DS-1811 are hence intended as averages (with related standard deviations). To complete our analysis, we performed a significance analysis using the well-known t-test. In the following, results in bold are intended as statistically significant ($p<0.0001$).
Results for DS-200 are not statistically validated since they presented no relevant variance.

\subsection{Test Set Classification Accuracy Results}

Table \ref{tab:results} shows the results for DS-200, DS-1811, and DS-1811-2.
Results on DS-200 achieved with the kNN classifier (for $k=5$) are comparable with those of GRAPSEC, although for different $k$ values ODSE obtains slightly inferior results.
Test set accuracy results obtained with ODSE using C-SVM operating in the embedding space equate those of GRAPSEC (same per-class errors). The kNN based reference system is systematically outperformed by the others, while the C-SVM classifier achieves a good result, however inferior to those obtained by ODSE and GRAPSEC.

Let us now focus on the results for DS-1811. ODSE systematically outperforms GRAPSEC and the kNN reference system, especially when using the C-SVM classifier in the dissimilarity space. Please note that differences here are statistically significant.
We note a more balanced per-class error distribution; in particular less errors are committed for the class of very soluble proteins (1), which proved to be hard to recognize also in our previous study \cite{grapsec_ijcnn_2013}.
Results obtained with the kNN operating directly in the input space are of poor quality, except for the $k=5$ case, where the system achieves competitive results.
It is worth noting that the results achieved with C-SVM operating in the input space are significantly better than any other system, although they are fairly comparable with the best result of ODSE. Standard deviations are in general low, demonstrating the stability of the results with respect to the different splits of the data.

Let us consider now the results on DS-1811-2, i.e., the split of DS-1811 with a perfectly balanced training set.
At first, we immediately recognize an overall decay of performance, regardless of the considered classifier. This is due to the fact that in DS-1811-2 the training set is made of randomly selected proteins (100 for each class), while in DS-1811 we suitably selected prototypes to better characterize the data.
Errors for each class are roughly the same, denoting a general weak sensitivity of the considered systems to the percentages of per-class patterns in the training set. However, please note that the training set of DS-1811 was not wildly imbalanced (70 soluble vs 110 insoluble proteins).
Finally, it is worth pointing out that ODSE configured with C-SVM achieves the best results, which are also statistically significant. From this we might conclude that ODSE is less sensitive to the specific training set instance adopted to train the model.
\begin{table}[tph!]\scriptsize
\caption{Test set classification accuracy results achieved on DS-200, DS-1811, and DS-1811-2.}
\begin{center}
\begin{tabular}{|c|c|c|c|c|c|}
\hline
\textbf{Class. System} & \textbf{Core Classifier} & \textbf{Params} & \textbf{\# Err. 0} & \textbf{\# Err. 1} & \textbf{Global TS Accuracy} \\
\hline
\multicolumn{6}{c}{\textbf{DS-200}} \\
\hline
\multirow{4}{*}{ODSE} & 
\multirow{3}{*}{\textit{k}-NN} & $k=1$ & 1 & 7 & 86.7\% \\
\cline{3-6}
 &  & $k=3$ & 1 & 6 & 88.4\% \\
\cline{3-6}
 &  & $k=5$ & 1 & 4 & 91.7\% \\
\cline{2-6}
& C-SVM & $C=2$ & 0 & 4 & 93.3\% \\
\hline
\multirow{3}{*}{GRAPSEC} & 
\multirow{3}{*}{\textit{k}-NN} & $k=1$ & 1 & 6 & 88.4\% \\
\cline{3-6}
 &  & $k=3$ & 1 & 4 & 91.7\% \\
\cline{3-6}
 &  & $k=5$ & 0 & 4 & 93.3\% \\
\hline
\multirow{3}{*}{kNN} & 
\multirow{3}{*}{-} &  $k=1$ & 24 & 0 & 60.0\% \\
\cline{3-6}
 &  & $k=3$ & 20 & 0 & 66.6\% \\
\cline{3-6}
 &  & $k=5$ & 24 & 0 & 66.0\% \\
\hline
C-SVM & - & $C=2$ & 2 & 5 & 88.3\% \\
\hline
\multicolumn{6}{c}{\textbf{DS-1811}} \\
\hline
\multirow{4}{*}{ODSE} & 
\multirow{3}{*}{\textit{k}-NN} & $k=1$ & 343.6 & 39.6 & 76.5\% ($\pm0.002$) \\
\cline{3-6}
& & $k=3$ & 306.6 & 36.0 & 78.9\% ($\pm0.001$) \\
\cline{3-6}
& & $k=5$ & 306.5 & 37.4 & 78.9\% ($\pm0.001$) \\
\cline{2-6}
& C-SVM & $C=2$ & 245.6 & 28.2 & 83.2\% ($\pm0.001$) \\
\hline
\multirow{3}{*}{GRAPSEC} & 
\multirow{3}{*}{\textit{k}-NN} & $k=1$ & 382.4 & 41.0 & 74.0\% ($\pm0.001$) \\
\cline{3-6}
& & $k=3$ & 375.0 & 40.6 & 74.5\% ($\pm0.001$) \\
\cline{3-6}
& & $k=5$ & 368.2 & 39.8 & 74.9\% ($\pm0.002$) \\
\hline
\multirow{3}{*}{kNN} & 
\multirow{3}{*}{-} & $k=1$ & 1228.2 & 7.4 & 24.5\% ($\pm0.001$) \\
\cline{3-6}
& & $k=3$ & 1315.6 & 1.2 & 19.4\% ($\pm0.002$) \\
\cline{3-6}
& & $k=5$ & 322.4 & 1.0 & 80.1\% ($\pm0.001$) \\
\hline
C-SVM & - & $C=2$ & 233.0 & 33.0 & \textbf{83.6}\% ($\pm0.002$) \\
\hline
\multicolumn{6}{c}{\textbf{DS-1811-2}} \\
\hline
\multirow{4}{*}{ODSE} & 
\multirow{3}{*}{\textit{k}-NN} & $k=1$ & 345.4 & 40.6 & 76.3\% ($\pm0.003$) \\
\cline{3-6}
& & $k=3$ & 308.6 & 37.0 & 78.8\% ($\pm0.003$) \\
\cline{3-6}
& & $k=5$ & 307.6 & 37.2 & 78.6\% ($\pm0.001$) \\
\cline{2-6}
& C-SVM & $C=2$ & 285.2 & 28.2 & \textbf{80.8}\% ($\pm0.025$) \\
\hline
\multirow{3}{*}{GRAPSEC} & 
\multirow{3}{*}{\textit{k}-NN} & $k=1$ & 394.6 & 45.8 & 73.0\% ($\pm0.002$) \\
\cline{3-6}
& & $k=3$ & 384.0 & 44.4 & 73.7\% ($\pm0.001$) \\
\cline{3-6}
& & $k=5$ & 372.2 & 42.2 & 74.6\% ($\pm0.003$) \\
\hline
\multirow{3}{*}{kNN} & 
\multirow{3}{*}{-} & $k=1$ & 1222.4 & 5.2 & 24.7\% ($\pm0.003$) \\
\cline{3-6}
& & $k=3$ & 1223.4 & 5.0 & 24.7\% ($\pm0.003$) \\
\cline{3-6}
& & $k=5$ & 420.6 & 5.6 & 73.9\% ($\pm0.008$) \\
\hline
C-SVM & - & $C=2$ & 362.6 & 16.2 & 76.8\% ($\pm0.001$) \\
\hline
\end{tabular}
\end{center}
\label{tab:results}
\end{table}

\section{Conclusions}
\label{sec:conclusions}

In this paper, we have evaluated the effectiveness and versatility of the ODSE classification system when processing sequences. Notably, we have considered an application dealing with the E. coli proteome classification. We focused on the recognition/discrimination of soluble and insoluble proteins, on the base of sequences of identifiers denoting amino acid residues.
Recognition of solubility/aggregation propensity of proteins is a very important research topic. In fact, aggregation of proteins is at the basis of many misfolding diseases, such as Parkinson and Alzheimer.
Each protein of our dataset was initially associated with a solubility degree and it was represented as a sequence of symbols, denoting the 20 amino acid residues.
Experiments presented in this paper have been carried out by considering different reference systems and several splits of the original data. The results obtained with ODSE denoted competitive test set classification accuracy percentages. Overall, the results presented in this paper strengthen the effectiveness of the ODSE system when dealing with structured data (sequences of symbols in this case). It is important to underline that the experiments have been performed without the need to tune any system component specifically for the considered problem (i.e., the E. coli solubility recognition), thus confirming that ODSE can be considered as a widely applicable classification system.

Future research directions involve the application of ODSE in other sequence-based pattern recognition problems, such as recognition of events in smart grid data. We also plan to cast the herein presented E. coli proteome analysis as a function approximation problem, i.e., by considering the continuous solubility degree as the target output signal.

\bibliographystyle{abbrvnat}
\bibliography{/home/lorenzo/University/Research/Publications/Bibliography.bib}

\begin{thebibliography}{49}
\providecommand{\natexlab}[1]{#1}
\providecommand{\url}[1]{\texttt{#1}}
\expandafter\ifx\csname urlstyle\endcsname\relax
  \providecommand{\doi}[1]{doi: #1}\else
  \providecommand{\doi}{doi: \begingroup \urlstyle{rm}\Url}\fi

\bibitem[Agostini et~al.(2012)Agostini, Vendruscolo, and
  Tartaglia]{Agostini2012237}
F.~Agostini, M.~Vendruscolo, and G.~G. Tartaglia.
\newblock {Sequence-Based Prediction of Protein Solubility}.
\newblock \emph{Journal of Molecular Biology}, 421\penalty0 (2-3):\penalty0
  237--241, 2012.
\newblock ISSN 0022-2836.
\newblock \doi{10.1016/j.jmb.2011.12.005}.

\bibitem[Batista et~al.(2010)Batista, Granger, and
  Sabourin]{batista+granger+sabourin2010}
L.~Batista, E.~Granger, and R.~Sabourin.
\newblock {Applying Dissimilarity Representation to Off-Line Signature
  Verification}.
\newblock In \emph{{Proceedings of the 2010 20th International Conference on
  Pattern Recognition}}, {ICPR '10}, pages 1293--1297, 2010.
\newblock ISBN 978-0-7695-4109-9.
\newblock \doi{10.1109/ICPR.2010.322}.

\bibitem[Bianchi et~al.(2014)Bianchi, Livi, Rizzi, and Sadeghian]{gralg_2012}
F.~M. Bianchi, L.~Livi, A.~Rizzi, and A.~Sadeghian.
\newblock A {G}ranular {C}omputing approach to the design of optimized graph
  classification systems.
\newblock \emph{Soft Computing}, 18\penalty0 (2):\penalty0 393--412, 2014.
\newblock ISSN 1432-7643.
\newblock \doi{10.1007/s00500-013-1065-z}.

\bibitem[Bianchi et~al.(2015)Bianchi, Livi, and Rizzi]{cagata}
F.~M. Bianchi, L.~Livi, and A.~Rizzi.
\newblock Two density-based k-means initialization algorithms for non-metric
  data clustering.
\newblock \emph{Pattern Analysis and Applications}, pages 1--19, 2015.
\newblock ISSN 1433-7541.
\newblock \doi{10.1007/s10044-014-0440-4}.

\bibitem[Bicego et~al.(2004)Bicego, Murino, and Figueiredo]{Bicego20042281}
M.~Bicego, V.~Murino, and M.~A.~T. Figueiredo.
\newblock {Similarity-based classification of sequences using hidden Markov
  models}.
\newblock \emph{Pattern Recognition}, 37\penalty0 (12):\penalty0 2281--2291,
  2004.
\newblock ISSN 0031-3203.
\newblock \doi{10.1016/j.patcog.2004.04.005}.

\bibitem[Cala{\~n}a et~al.(2010)Cala{\~n}a, Reyes, Alzate, and Duin]{5597597}
Y.~Cala{\~n}a, E.~Reyes, M.~Alzate, and R.~P.~W. Duin.
\newblock {Prototype Selection for Dissimilarity Representation by a Genetic
  Algorithm}.
\newblock In \emph{{Proceedings of the 20th International Conference on Pattern
  Recognition}}, pages 177--180, Aug. 2010.
\newblock \doi{10.1109/ICPR.2010.52}.

\bibitem[{Del Vescovo} and Rizzi(2007{\natexlab{a}})]{delvescovo+rizzi2007}
G.~{Del Vescovo} and A.~Rizzi.
\newblock {Automatic Classification of Graphs by Symbolic Histograms}.
\newblock In \emph{{Proceedings of the 2007 IEEE International Conference on
  Granular Computing}}, {GRC '07}, pages 410--416. IEEE Computer Society,
  2007{\natexlab{a}}.
\newblock ISBN 0-7695-3032-X.
\newblock \doi{10.1109/GRC.2007.46}.

\bibitem[{Del Vescovo} and
  Rizzi(2007{\natexlab{b}})]{delvescovo07OnlineHandwritingRecognitionbySymbolicHistogramsApproach}
G.~{Del Vescovo} and A.~Rizzi.
\newblock {Online Handwriting Recognition by the Symbolic Histograms Approach}.
\newblock In \emph{{Proceedings of the 2007 IEEE International Conference on
  Granular Computing}}, {GRC '07}, pages 686--700, Washington, DC, USA,
  2007{\natexlab{b}}. IEEE Computer Society.
\newblock ISBN 0-7695-3032-X.
\newblock \doi{10.1109/GRC.2007.116}.

\bibitem[Dill and MacCallum(2012)]{dill2012protein}
K.~A. Dill and J.~L. MacCallum.
\newblock {The protein-folding problem, 50 years on}.
\newblock \emph{Science}, 338\penalty0 (6110):\penalty0 1042--1046, 2012.

\bibitem[Duin and P\c{e}kalska(2012)]{Duin2012826}
R.~P.~W. Duin and E.~P\c{e}kalska.
\newblock {The dissimilarity space: Bridging structural and statistical pattern
  recognition}.
\newblock \emph{Pattern Recognition Letters}, 33\penalty0 (7):\penalty0
  826--832, 2012.
\newblock ISSN 0167-8655.
\newblock \doi{10.1016/j.patrec.2011.04.019}.

\bibitem[Frauenfelder and Wolynes(1994)]{frauenfelder1994}
H.~Frauenfelder and P.~Wolynes.
\newblock {Proteins: where physics of simplicity and complexity meet}.
\newblock \emph{Physics Today}, 47:\penalty0 58--65, 1994.

\bibitem[Gabrys and Bargiela(2000)]{Gabrys2000}
B.~Gabrys and A.~Bargiela.
\newblock {General Fuzzy Min-Max Neural Network for Clustering and
  Classification}.
\newblock \emph{IEEE Transactions on Neural Networks}, 11\penalty0
  (3):\penalty0 769--783, 2000.

\bibitem[Giuliani et~al.(2002)Giuliani, Benigni, Zbilut, {Webber Jr.},
  Sirabella, and Colosimo]{giuliani2002}
A.~Giuliani, R.~Benigni, J.~P. Zbilut, C.~L. {Webber Jr.}, P.~Sirabella, and
  A.~Colosimo.
\newblock {Nonlinear Signal Analysis Methods in the Elucidation of Protein
  Sequence---Structure Relationships}.
\newblock \emph{ChemInform}, 33\penalty0 (28):\penalty0 1471--1492, 2002.
\newblock ISSN 1522-2667.
\newblock \doi{10.1002/chin.200228300}.

\bibitem[{Hero III} et~al.(2002){Hero III}, Ma, Michel, and
  Gorman]{gs:HeroEtAl2002}
A.~O. {Hero III}, B.~Ma, O.~J.~J. Michel, and J.~Gorman.
\newblock {Applications of entropic spanning graphs}.
\newblock \emph{IEEE Signal Processing Magazine}, 19\penalty0 (5):\penalty0
  85--95, Sept. 2002.
\newblock \doi{10.1109/MSP.2002.1028355}.

\bibitem[Iosifidis et~al.(2012)Iosifidis, Tefas, and
  Pitas]{10.1109/TKDE.2012.223}
A.~Iosifidis, A.~Tefas, and I.~Pitas.
\newblock {Multidimensional Sequence Classification based on Fuzzy Distances
  and Discriminant Analysis}.
\newblock \emph{IEEE Transactions on Knowledge and Data Engineering},
  99\penalty0 (PrePrints):\penalty0 1, 2012.
\newblock ISSN 1041-4347.
\newblock \doi{10.1109/TKDE.2012.223}.

\bibitem[Jacobs et~al.(2000)Jacobs, Weinshall, and
  Gdalyahu]{Jacobs:2000:CND:355091.355095}
D.~W. Jacobs, D.~Weinshall, and Y.~Gdalyahu.
\newblock {Classification with Nonmetric Distances: Image Retrieval and Class
  Representation}.
\newblock \emph{IEEE Transactions on Pattern Analysis and Machine
  Intelligence}, 22\penalty0 (6):\penalty0 583--600, June 2000.
\newblock ISSN 0162-8828.
\newblock \doi{10.1109/34.862197}.

\bibitem[Lan et~al.(2009)Lan, Tan, Su, and Lu]{4509437}
M.~Lan, C.~L. Tan, J.~Su, and Y.~Lu.
\newblock {Supervised and Traditional Term Weighting Methods for Automatic Text
  Categorization}.
\newblock \emph{IEEE Transactions on Pattern Analysis and Machine
  Intelligence}, 31\penalty0 (4):\penalty0 721--735, Apr 2009.
\newblock ISSN 0162-8828.
\newblock \doi{10.1109/TPAMI.2008.110}.

\bibitem[Livi and Rizzi(2013)]{gm_survey}
L.~Livi and A.~Rizzi.
\newblock The graph matching problem.
\newblock \emph{Pattern Analysis and Applications}, 16\penalty0 (3):\penalty0
  253--283, 2013.
\newblock ISSN 1433-7541.
\newblock \doi{10.1007/s10044-012-0284-8}.

\bibitem[Livi et~al.(2012)Livi, {Del Vescovo}, and
  Rizzi]{livi+delvescovo+rizzi_seriation+gradis}
L.~Livi, G.~{Del Vescovo}, and A.~Rizzi.
\newblock Graph recognition by seriation and frequent substructures mining.
\newblock In \emph{{Proceedings of the First International Conference on
  Pattern Recognition Applications and Methods}}, volume~1, pages 186--191,
  Feb. 2012.
\newblock ISBN 978-989-8425-98-0.
\newblock \doi{10.5220/0003733201860191}.

\bibitem[Livi et~al.(2013{\natexlab{a}})Livi, Bianchi, Rizzi, and
  Sadeghian]{odse2_ijcnn_2013}
L.~Livi, F.~M. Bianchi, A.~Rizzi, and A.~Sadeghian.
\newblock Dissimilarity space embedding of labeled graphs by a clustering-based
  compression procedure.
\newblock In \emph{{Proceedings of the 2013 International Joint Conference on
  Neural Networks}}, pages 1646--1653, Aug 2013{\natexlab{a}}.
\newblock ISBN 978-1-4673-6129-3.
\newblock \doi{10.1109/IJCNN.2013.6706937}.

\bibitem[Livi et~al.(2013{\natexlab{b}})Livi, {Del Vescovo}, and
  Rizzi]{seriation+gradis_lncs_2012}
L.~Livi, G.~{Del Vescovo}, and A.~Rizzi.
\newblock Combining graph seriation and substructures mining for graph
  recognition.
\newblock In P.~{Latorre Carmona}, J.~S. S{\'a}nchez, and A.~L.~N. Fred,
  editors, \emph{{Pattern Recognition - Applications and Methods}}, volume 204
  of \emph{{Advances in Intelligent and Soft Computing}}, pages 79--91.
  Springer Berlin Heidelberg, 2013{\natexlab{b}}.
\newblock ISBN 978-3-642-36529-4.
\newblock \doi{10.1007/978-3-642-36530-0_7}.

\bibitem[Livi et~al.(2013{\natexlab{c}})Livi, Tahayori, Sadeghian, and
  Rizzi]{t2apdiss__ifsanafips2013}
L.~Livi, H.~Tahayori, A.~Sadeghian, and A.~Rizzi.
\newblock Aggregating $\alpha$-planes for type-2 fuzzy set matching.
\newblock In \emph{{2013 Joint IFSA World Congress and NAFIPS Annual Meeting
  (IFSA/NAFIPS)}}, pages 860--865, 2013{\natexlab{c}}.
\newblock \doi{10.1109/IFSA-NAFIPS.2013.6608513}.

\bibitem[Livi et~al.(2014{\natexlab{a}})Livi, {Del Vescovo}, Rizzi, and
  {Frattale Mascioli}]{spare_graph_2013}
L.~Livi, G.~{Del Vescovo}, A.~Rizzi, and F.~M. {Frattale Mascioli}.
\newblock Building pattern recognition applications with the {SPARE} library.
\newblock \emph{ArXiv preprint arXiv:1410.5263}, Oct 2014{\natexlab{a}}.

\bibitem[Livi et~al.(2014{\natexlab{b}})Livi, Rizzi, and Sadeghian]{odse}
L.~Livi, A.~Rizzi, and A.~Sadeghian.
\newblock Optimized dissimilarity space embedding for labeled graphs.
\newblock \emph{Information Sciences}, 266:\penalty0 47--64,
  2014{\natexlab{b}}.
\newblock ISSN 0020-0255.
\newblock \doi{10.1016/j.ins.2014.01.005}.

\bibitem[Livi et~al.(2014{\natexlab{c}})Livi, Rizzi, and
  Sadeghian]{odse2__arxiv}
L.~Livi, A.~Rizzi, and A.~Sadeghian.
\newblock Designing labeled graph classifiers by exploiting the {R}{\'e}nyi
  entropy of the dissimilarity representation.
\newblock \emph{ArXiv preprint arXiv:1408.5286}, Aug 2014{\natexlab{c}}.

\bibitem[Livi et~al.(2014{\natexlab{d}})Livi, Tahayori, Sadeghian, and
  Rizzi]{it2fs_matching}
L.~Livi, H.~Tahayori, A.~Sadeghian, and A.~Rizzi.
\newblock {Distinguishability of interval type-2 fuzzy sets data by analyzing
  upper and lower membership functions}.
\newblock \emph{Applied Soft Computing}, 17:\penalty0 79--89,
  2014{\natexlab{d}}.
\newblock ISSN 1568-4946.
\newblock \doi{10.1016/j.asoc.2013.12.020}.

\bibitem[Melin and Castillo(2013)]{melin2013review}
P.~Melin and O.~Castillo.
\newblock A review on the applications of type-2 fuzzy logic in classification
  and pattern recognition.
\newblock \emph{Expert Systems with Applications}, 40\penalty0 (13):\penalty0
  5413--5423, 2013.

\bibitem[Melin and Castillo(2014)]{melin2014review}
P.~Melin and O.~Castillo.
\newblock A review on type-2 fuzzy logic applications in clustering,
  classification and pattern recognition.
\newblock \emph{Applied Soft Computing}, 21:\penalty0 568--577, 2014.

\bibitem[Melin et~al.(2014)Melin, Amezcua, Valdez, and Castillo]{melin2014new}
P.~Melin, J.~Amezcua, F.~Valdez, and O.~Castillo.
\newblock A new neural network model based on the lvq algorithm for multi-class
  classification of arrhythmias.
\newblock \emph{Information Sciences}, 279:\penalty0 483--497, 2014.

\bibitem[Moharrer et~al.(2015)Moharrer, Tahayori, Livi, Sadeghian, and
  Rizzi]{usit2_2012}
M.~Moharrer, H.~Tahayori, L.~Livi, A.~Sadeghian, and A.~Rizzi.
\newblock Interval type-2 fuzzy sets to model linguistic label perception in
  online services satisfaction.
\newblock \emph{Soft Computing}, 19\penalty0 (1):\penalty0 237--250, 2015.
\newblock ISSN 1432-7643.
\newblock \doi{10.1007/s00500-014-1246-4}.

\bibitem[Niwa et~al.(2009)Niwa, Ying, Saito, Jin, Takada, Ueda, and
  Taguchi]{niwa2009}
T.~Niwa, B.-W. Ying, K.~Saito, W.~Jin, S.~Takada, T.~Ueda, and H.~Taguchi.
\newblock {Bimodal protein solubility distribution revealed by an aggregation
  analysis of the entire ensemble of Escherichia coli proteins}.
\newblock \emph{Proceedings of the National Academy of Sciences}, 106\penalty0
  (11):\penalty0 4201--4206, 2009.
\newblock \doi{10.1073/pnas.0811922106}.

\bibitem[Niwa et~al.(2012)Niwa, Kanamori, Ueda, and Taguchi]{Niwa05062012}
T.~Niwa, T.~Kanamori, T.~Ueda, and H.~Taguchi.
\newblock {Global analysis of chaperone effects using a reconstituted cell-free
  translation system}.
\newblock \emph{Proceedings of the National Academy of Sciences}, 109\penalty0
  (23):\penalty0 8937--8942, 2012.
\newblock \doi{10.1073/pnas.1201380109}.

\bibitem[Pedrycz et~al.(2014)Pedrycz, Lu, Liu, Wang, and
  Wang]{pedrycz_timeseries}
W.~Pedrycz, W.~Lu, X.~Liu, W.~Wang, and L.~Wang.
\newblock {Human-centric analysis and interpretation of time series: a
  perspective of granular computing}.
\newblock \emph{Soft Computing}, pages 1--15, 2014.
\newblock ISSN 1432-7643.
\newblock \doi{10.1007/s00500-013-1213-5}.

\bibitem[Pr{\'i}ncipe(2010)]{principe2010}
J.~C. Pr{\'i}ncipe.
\newblock \emph{{Information Theoretic Learning: Renyi's Entropy and Kernel
  Perspectives}}.
\newblock {Information Science and Statistics}. Springer, 2010.
\newblock ISBN 9781441915696.

\bibitem[Riesen and Bunke(2010)]{riesen+bunke2010}
K.~Riesen and H.~Bunke.
\newblock \emph{{Graph Classification and Clustering Based on Vector Space
  Embedding}}.
\newblock {Series in Machine Perception and Artificial Intelligence}. World
  Scientific Pub Co Inc, 2010.
\newblock ISBN 9789814304719.

\bibitem[Rizzi et~al.(2002)Rizzi, Panella, and {Frattale Mascioli}]{rizzi2002}
A.~Rizzi, M.~Panella, and F.~M. {Frattale Mascioli}.
\newblock {Adaptive resolution min-max classifiers}.
\newblock \emph{IEEE Transactions on Neural Networks}, 13:\penalty0 402--414,
  Mar. 2002.
\newblock ISSN 1045-9227.

\bibitem[Rizzi et~al.(2012)Rizzi, {Del Vescovo}, Livi, and {Frattale
  Mascioli}]{delvescovo_rlgradis_2012}
A.~Rizzi, G.~{Del Vescovo}, L.~Livi, and F.~M. {Frattale Mascioli}.
\newblock A new granular computing approach for sequences representation and
  classification.
\newblock In \emph{{Proceedings of the 2012 International Joint Conference on
  Neural Networks}}, pages 2268--2275, June 2012.
\newblock ISBN 978-1-4673-1489-3.
\newblock \doi{10.1109/IJCNN.2012.6252680}.

\bibitem[Rizzi et~al.(2013{\natexlab{a}})Rizzi, Livi, Tahayori, and
  Sadeghian]{t2vsdiss__ifsanafips2013}
A.~Rizzi, L.~Livi, H.~Tahayori, and A.~Sadeghian.
\newblock Matching general type-2 fuzzy sets by comparing the vertical slices.
\newblock In \emph{{2013 Joint IFSA World Congress and NAFIPS Annual Meeting
  (IFSA/NAFIPS)}}, pages 866--871, 2013{\natexlab{a}}.
\newblock \doi{10.1109/IFSA-NAFIPS.2013.6608514}.

\bibitem[Rizzi et~al.(2013{\natexlab{b}})Rizzi, Possemato, Livi, Sebastiani,
  Giuliani, and {Frattale Mascioli}]{grapsec_ijcnn_2013}
A.~Rizzi, F.~Possemato, L.~Livi, A.~Sebastiani, A.~Giuliani, and F.~M.
  {Frattale Mascioli}.
\newblock A dissimilarity-based classifier for generalized sequences by a
  {G}ranular {C}omputing approach.
\newblock In \emph{{Proceedings of the 2013 International Joint Conference on
  Neural Networks}}, pages 2397--2404, Aug 2013{\natexlab{b}}.
\newblock ISBN 978-1-4673-6129-3.
\newblock \doi{10.1109/IJCNN.2013.6707041}.

\bibitem[Sadeghian(2001)]{1009136}
A.~Sadeghian.
\newblock {Nonlinear neuro-fuzzy prediction: methodology, design and
  applications}.
\newblock In \emph{{The 10th IEEE International Conference on Fuzzy Systems}},
  volume~3, pages 1022--1026, 2001.
\newblock \doi{10.1109/FUZZ.2001.1009136}.

\bibitem[Samak et~al.(2012)Samak, Gunter, and Wang]{6404416}
T.~Samak, D.~Gunter, and Z.~Wang.
\newblock Prediction of protein solubility in {E}. coli.
\newblock In \emph{2012 IEEE 8th International Conference on E-Science
  (e-Science)}, pages 1--8, Oct 2012.
\newblock \doi{10.1109/eScience.2012.6404416}.

\bibitem[S{\'a}nchez and Melin(2014)]{sanchez2014optimization}
D.~S{\'a}nchez and P.~Melin.
\newblock Optimization of modular granular neural networks using hierarchical
  genetic algorithms for human recognition using the ear biometric measure.
\newblock \emph{Engineering Applications of Artificial Intelligence},
  27:\penalty0 41--56, 2014.

\bibitem[Shtilerman et~al.(1999)Shtilerman, Lorimer, and {Walter
  Englander}]{shtilerman1999}
M.~Shtilerman, G.~H. Lorimer, and S.~{Walter Englander}.
\newblock {Chaperonin Function: Folding by Forced Unfolding}.
\newblock \emph{Science}, 284\penalty0 (5415):\penalty0 822--825, 1999.
\newblock \doi{10.1126/science.284.5415.822}.

\bibitem[Smialowski et~al.(2007)Smialowski, Martin-Galiano, Mikolajka,
  Girschick, Holak, and Frishman]{smialowski2007protein}
P.~Smialowski, A.~J. Martin-Galiano, A.~Mikolajka, T.~Girschick, T.~A. Holak,
  and D.~Frishman.
\newblock Protein solubility: sequence based prediction and experimental
  verification.
\newblock \emph{Bioinformatics}, 23\penalty0 (19):\penalty0 2536--2542, 2007.

\bibitem[Tappert et~al.(1990)Tappert, Suen, and Wakahara]{10.1109/34.57669}
C.~Tappert, C.~Suen, and T.~Wakahara.
\newblock {The State of the Art in Online Handwriting Recognition}.
\newblock \emph{IEEE Transactions on Pattern Analysis and Machine
  Intelligence}, 12\penalty0 (8):\penalty0 787--808, 1990.
\newblock ISSN 0162-8828.
\newblock \doi{10.1109/34.57669}.

\bibitem[Taubes(1996)]{Taubes15031996}
G.~Taubes.
\newblock {Protein Chemistry: Misfolding the Way to Disease}.
\newblock \emph{Science}, 271\penalty0 (5255):\penalty0 1493--1495, 1996.
\newblock \doi{10.1126/science.271.5255.1493}.

\bibitem[Trujillo-Pulgar{\'i}n and Orozco-Alzate(2013)]{trujillo2013parzen}
C.~Trujillo-Pulgar{\'i}n and M.~Orozco-Alzate.
\newblock {Parzen classification in generalised dissimilarity spaces}.
\newblock \emph{Electronics Letters}, 49\penalty0 (3):\penalty0 192--193, 2013.

\bibitem[Xiaohui et~al.(2014)Xiaohui, Feng, Xuehai, Jingbo, and
  Nana]{xiaohui2014predicting}
N.~Xiaohui, S.~Feng, H.~Xuehai, X.~Jingbo, and L.~Nana.
\newblock Predicting the protein solubility by integrating chaos games
  representation and entropy in information theory.
\newblock \emph{Expert Systems with Applications}, 41\penalty0 (4):\penalty0
  1672--1679, 2014.

\bibitem[Xing et~al.(2010)Xing, Pei, and Keogh]{xing2010}
Z.~Xing, J.~Pei, and E.~Keogh.
\newblock {A brief survey on sequence classification}.
\newblock \emph{SIGKDD Explorations Newsletter}, 12:\penalty0 40--48, Nov.
  2010.
\newblock ISSN 1931-0145.
\newblock \doi{10.1145/1882471.1882478}.

\end{thebibliography}
\end{document}